\documentclass[sigconf]{acmart}
\usepackage{graphicx}
\usepackage{booktabs}
\usepackage[ruled]{algorithm2e}
\usepackage{multirow}
\usepackage{colortbl}
\usepackage{wrapfig}
\usepackage{enumitem}
\usepackage{url}

\AtBeginDocument{%
  \providecommand\BibTeX{{%
    \normalfont B\kern-0.5em{\scshape i\kern-0.25em b}\kern-0.8em\TeX}}}

\copyrightyear{2023}
\acmYear{2023}
\setcopyright{acmlicensed}\acmConference[KDD '23]{Proceedings of the 29th ACM SIGKDD Conference on Knowledge Discovery and Data Mining}{August 6--10, 2023}{Long Beach, CA, USA}
\acmBooktitle{Proceedings of the 29th ACM SIGKDD Conference on Knowledge Discovery and Data Mining (KDD '23), August 6--10, 2023, Long Beach, CA, USA}
\acmPrice{15.00}
\acmDOI{10.1145/3580305.3599460}
\acmISBN{979-8-4007-0103-0/23/08}

\begin{document}

\title{Partial-label Learning with Mixed Closed-set and Open-set Out-of-candidate Examples}

\author{Shuo He}
\affiliation{%
 \institution{University of Electronic Science and Technology of China}
 \city{Chengdu}
 \country{China}
}
\email{shuohe@std.uestc.edu.cn}

\author{Lei Feng}
\affiliation{%
 \institution{Nanyang Technological University}
 \city{Singapore}
 \country{Singapore}
}
\email{lfengqaq@gmail.com}

\author{Guowu Yang}
\authornote{Corresponding author}
\affiliation{%
 \institution{University of Electronic Science and Technology of China}
 \city{Chengdu}
 \country{China}
}
\email{guowu@uestc.edu.cn}

\renewcommand{\shortauthors}{Shuo He, Lei Feng, \& Guowu Yang}

\begin{abstract}
Partial-label learning (PLL) relies on a key assumption that the true label of each training example must be in the candidate label set. This restrictive assumption may be violated in complex real-world scenarios, and thus the true label of some collected examples could be unexpectedly outside the assigned candidate label set. In this paper, we term the examples whose true label is outside the candidate label set OOC (\underline{o}ut-\underline{o}f-\underline{c}andidate) examples, and pioneer a new PLL study to learn with OOC examples.
We consider two types of OOC examples in reality, i.e., the closed-set/open-set OOC examples whose true label is inside/outside the known label space. To solve this new PLL problem, we first calculate the \emph{wooden cross-entropy loss} from candidate and non-candidate labels respectively, and dynamically differentiate the two types of OOC examples based on specially designed criteria. Then, for closed-set OOC examples, we conduct reversed label disambiguation in the non-candidate label set; for open-set OOC examples, we leverage them for training by utilizing an effective regularization strategy that dynamically assigns random candidate labels from the candidate label set. In this way, the two types of OOC examples can be differentiated and further leveraged for model training. Extensive experiments demonstrate that our proposed method outperforms state-of-the-art PLL methods.
\end{abstract}



\maketitle

\section{Introduction}
Modern annotation-hungry deep learning techniques have a serious need for large amounts of correctly labeled data. However, perfectly labeled data in real-world scenarios is considerably scarce due to the difficulty in data annotation \cite{zhou2018brief,li2019towards,ahn2019weakly}. To overcome this obstacle, partial-label learning (PLL), which serves as a labeling-friendly weakly-supervised learning paradigm, has alleviated the demand of labeling only one ground-truth label and allows to assign a candidate label set instead. PLL relies on a key assumption that the true label of each training example must be in the assigned candidate label set. However, this restrictive assumption may be violated in complex real-world scenarios \cite{xia2022extended}, and thus the true label of some collected examples could be unexpectedly outside the assigned candidate label set. Such examples are termed OOC (\underline{o}ut-\underline{o}f-\underline{c}andidate) examples in this paper whose true label is outside the candidate label set. Furthermore, due to the open and dynamic nature of the real world, it is unrealistic to expect an annotator can define all classes in the real world and has the corresponding domain knowledge to identify them perfectly. Therefore, one may encounter two types of OOC examples in reality: the closed-set/open-set OOC example whose true label is inside/outside the known label space. 
\begin{figure}[!t]
\centering
\includegraphics[width=0.45\textwidth]{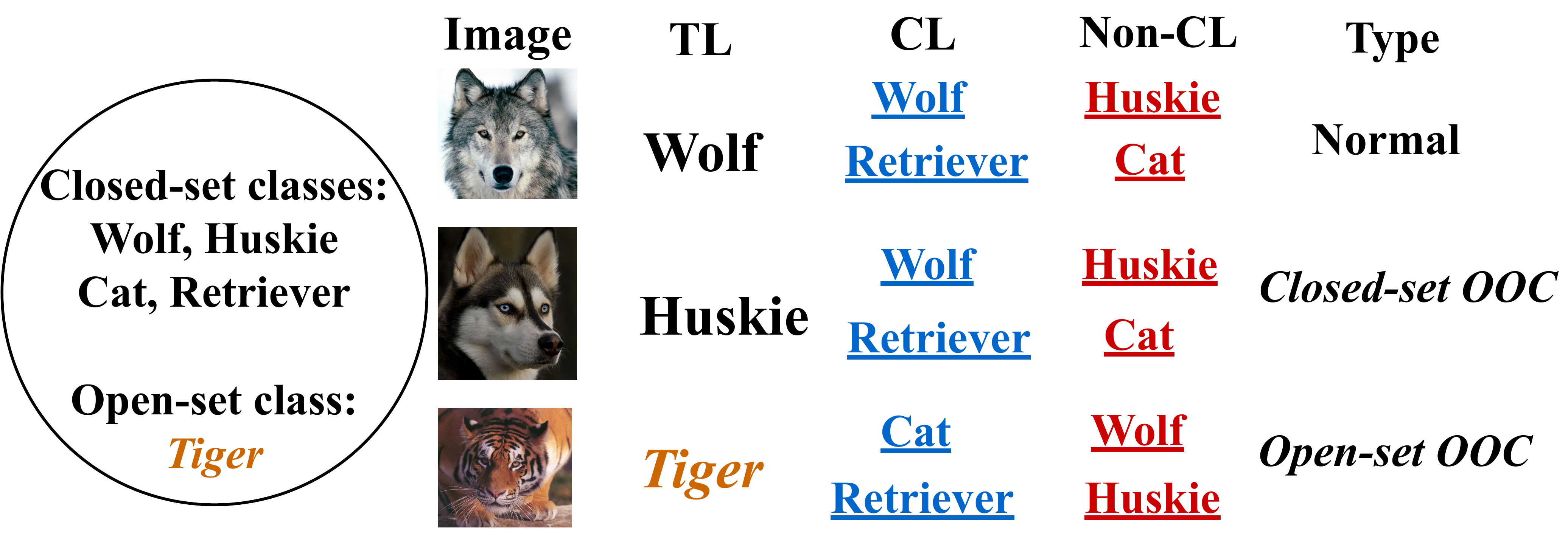} 
	\caption{An example of three different types of partially-labeled images. TL, CLs, and Non-CL denote ``True Label", ``Candidate Labels" and ``Non-candidate Labels", respectively.}
	\label{OOC_illustration}
\end{figure}

Here, we provide an intuitive illustration of OOC examples in real-world scenarios. As shown in Figure \ref{OOC_illustration}, suppose that defined domain classes only contain \textit{Wolf}, \textit{Cat}, \textit{Retriever} and \textit{Huskie}, and a non-expert annotator has a little knowledge about the first three classes, and completely lacks the knowledge about \textit{Huskie}. The annotator needs to label these three images from the open world. For the first image, he hesitates to choose between \textit{Wolf} and \textit{Retriever} due to similar semantics, and thus assigns a candidate label set (\textit{Wolf}, \textit{Retriever}), while he certainly excludes \textit{Cat} and is ignorant of \textit{Huskie}, thereby forming a corresponding non-candidate label set (\textit{Cat}, \textit{Huskie}), which is a normal partially-labeled example. For the second image, due to the lack of knowledge about \textit{Huskie}, he misses it regrettably and chooses two familiar domain classes (\textit{Wolf}, \textit{Retriever}) instead, which is the so-called closed-set OOC example. For the third image, since \textit{Tiger} is an open-set class unseen in defined domain classes, he fails to identify the true label inevitably, and has to choose several visual-similar domain classes (\textit{Cat}, \textit{Retriever}), which is so-called open-set OOC example. 

In this paper, we pioneer a new PLL study to learn from such two types of OOC examples. To solve this new PLL problem, we propose a unified framework including three synergistic parts: OOC selection, (reversed) label disambiguation, and random candidate generation. In the first part, the core is to discriminate two types of OOC examples effectively. For this purpose, we first point out the weakness of ordinary cross-entropy (CE) loss and then utilize the \textit{wooden} CE loss which only considers the minimum CE loss. Based on this, we further calculate the per-example wooden CE loss from candidate and non-candidate labels respectively and dynamically differentiate two types of OOC examples based on specially designed criteria. After this, instead of discarding these selected OOC examples, we further leverage them for model training, as they also contain useful information for generalization. Specifically, for closed-set OOC examples, we propose \emph{reversed label disambiguation} that identifies the unknown true label only in the non-candidate label set; for open-set OOC examples, we use them for training by utilizing an effective regularization strategy that dynamically assigns random candidate labels from the candidate label set. In this way, the two types of OOC examples can be differentiated and further leveraged for model training. Extensive experiments demonstrate that our proposed method outperforms state-of-the-art methods. Our main contributions are summarized as follows:      
\begin{itemize}[leftmargin=0.4cm]
\item To the best of our knowledge, we pioneer a new PLL study to learn from mixed closed-set and open-set OOC examples simultaneously, and accordingly propose a novel framework to solve this problem. 
\item For OOC selection, we point out the weakness of partial-level CE loss and utilize the \emph{wooden} CE loss instead. Furthermore, we propose \emph{reversed label disambiguation} and \emph{random candidate generation} to leverage selected OOC examples for model training.
\item Empirically, extensive experiments show the superiority and effectiveness of our proposed method.
\end{itemize}

\begin{figure*}[!t]
\centering
\includegraphics[width=0.9\linewidth]{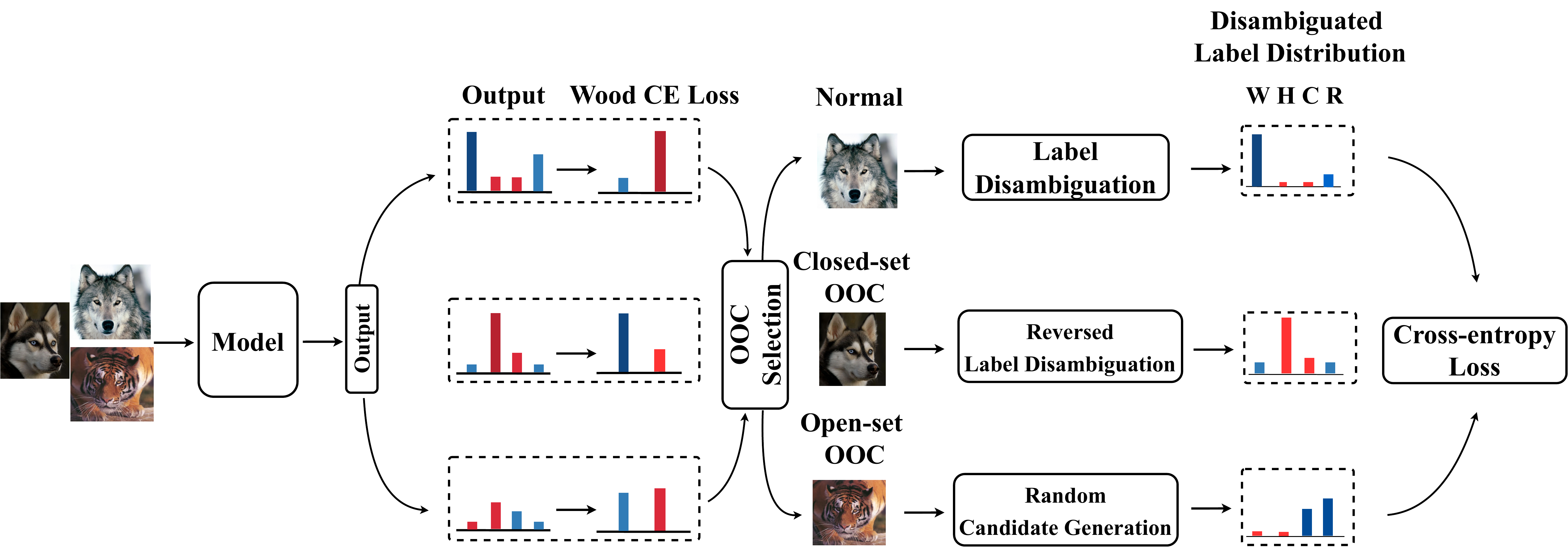} 
	\caption{Our proposed method first calculates the per-example wooden CE loss from both candidate and non-candidate labels respectively. Based on this, OOC selection differentiates closed-set and open-set OOC examples respectively based on specially designed criteria. Then, (reversed) label disambiguation is performed for normal (closed-set OOC) examples, and random candidate generation is conducted only for open-set OOC examples. The blue (red) color denotes the candidate (non-candidate) part. We abbreviate ``Wolf", ``Huskie", ``Cat", ``Retriever" to W, H, C, R.}
	\label{framework}
\end{figure*}

\section{Related Work}
In this section, we first review partial-label learning and then introduce open-set learning including three related tasks: open-set and open-world recognition, and open-set label-noise learning. 
\subsection{Partial-Label Learning}
Partial-label learning (PLL) deals with the problem where each training example is associated with a candidate label set, among which only the unknown one is the true label. Conventional PLL methods aim to induce a multi-class classifier based on hand-crafted features and given candidate label sets \cite{feng2019partial,tang2017confidence,xu2021instance,lyu2020self}, mainly including the average-based \cite{cour2011learning,hullermeier2006learning} and identification-based \cite{zhang2015solving,zhang2016partial,xu2019partial,wang2021adaptive,lyu2019gm,yan2020partial,dong2023aaai} frameworks. Recently, to scale to large-scale datasets, deep PLL methods have attracted much attention, which aims to end-to-end train a deep neural network from scratch with partially-labeled data \cite{wu2022revisiting,wang2022partial,zhang2021exploiting,wang2022solar,gong2022partial,yu2022partial,li2022active}. The first algorithms with the guarantee of risk and classifier consistency were proposed by \cite{feng2020provably,lv2020progressive}. Similarly, \cite{wen2021leveraged} proposed the leveraged weighted loss, a family of loss functions with risk consistency, that considered the trade-off between losses on candidate labels and non-candidate ones. Especially, \cite{wang2022pico} utilized contrastive representation learning and achieved state-of-the-art performances in PLL. 

\subsection{Open-set Learning}
Due to the open and dynamic nature of real-world \cite{shao2022open,zheng2022towards,chen2020learning,kong2021opengan,saito2021openmatch}, it is unrealistic to expect that one can define and identify all classes, and thereby supervise the ideal machine learning models. Therefore, training and test data may contain open-set examples whose class is outside the set of predefined training classes. There are several related tasks including open-set and open-world recognition and open-set noise learning. Particularly, the first two tasks focus on test-time open-set examples, while the third one concerns training-time open-set examples.    

\noindent\textbf{Open-set and open-world recognition.} Open-set recognition \cite{2020Recent,lu2022pmal,perera2020generative,oza2019c2ae,liu2020few,luo2020progressive} considers the scenario where unknown classes can appear during testing, and the model needs to recognize and reject examples of unknown classes. To achieve this goal, many methods have been proposed such as conventional methods based on support vector machines \cite{scheirer2012toward,jain2014multi} and distance ratio \cite{PRM2017Nearest}, and deep learning methods \cite{2016Towards, zhou2021learning}. Compared with open-set recognition, open-world recognition \cite{boult2019learning,cao2021open} not only needs to recognize open-set classes but also aims to learn and extend the set of known classes with them incrementally.      

\noindent\textbf{Open-set label-noise learning.} Different from the aforementioned two tasks, open-set label-noise learning deals with the problem where training data contains open-set noisy examples whose true label is outside the set of known training classes \cite{wu2021ngc}. To solve this problem, \cite{wang2018iterative} proposed an iterative learning framework that detects open-set noisy examples and reduces their weights in the training process. Furthermore, \cite{sachdeva2021evidentialmix,xia2022extended} considered the scenario where training data contained both open-set and closed-set noisy examples, and solved this problem with selection-based and transition-based methods respectively. In general, they always have a prejudice against open-set noisy examples, and aim to filter them out unhesitatingly. Recently, \cite{wei2021open} revealed a counter-intuitive observation that open-set examples can even benefit the robustness of deep neural networks against noisy labels, which inspired us to rethink the effect of open-set examples.     

\section{Problem Setup}
In this section, we introduce the necessary symbol and terminology to define the problem of partial-label learning (PLL) with mixed closed-set and open-set OOC (\underline{o}ut-\underline{o}f-\underline{c}andidate) examples.

Supposed that we have a partially-labeled dataset with $n$ examples ${\mathcal{D}}= \{ x_{i},{Y}_{i} \}_{i=1}^{n}$, each pair composes of an example $x_{i} \in \mathcal{X}$ and a corresponding candidate label set ${Y}_{i} \subset \mathcal{Y}$ where $\mathcal{X}$ and $\mathcal{Y} = \{ 1,2,...,c \}$ denote the input space and the label space respectively. Additionally, the non-candidate label set is denoted by $\widebar{Y}_{i} \subset \mathcal{Y}$. Also, each example $x_{i}$ associates an one-hot style label vector $\mathbf{y}_{i}$ where $y_{i}^{j} = 1 (0)$ means that the $j$-th label is a candidate (non-candidate) label. Similarly, we define a non-candidate label vector $\widebar{\mathbf{y}}$ where $\widebar{y}^{j}=1 (0)$ means the $j-$th label is a non-candidate (candidate) one. Particularly, there is a key assumption \cite{feng2020provably} in PLL that the unknown true label $y_i$ must be in the candidate label set ${Y}_{i}$ (i.e., $P(y_i \in {Y}_{i}|x_i,{Y}_{i})=1$) and not in the non-candidate label set (i.e., $P(y_i \notin \widebar{Y}_{i}|x_i,\widebar{Y}_{i})=1$). As mentioned before, such a restrictive assumption may be violated in complex real-world scenarios. Therefore, we consider a certain proportion ($\tau_{1}$ and $\tau_{2}$) of closed-set OOC examples $\mathcal{D}_{C}=\{ x_{i}, y_{i} \}_{i=1}^{\tau_{1} \times n}$ where $ y_{i} \notin Y_{i}$ and $ y_{i} \in \widebar{Y}_{i} $ and open-set OOC examples $\mathcal{D}_{O}=\{ \widetilde{x}_{i},y_{i} \}_{i=1}^{\tau_{2} \times n}$ where $\widetilde{x}_{i} \in \mathcal{X}$ and $ y_{i} \notin \mathcal{Y} $. In this work, we pioneer a new PLL study to learn from such mixed closed-set OOC examples $\mathcal{D}_{C}$ and open-set OOC examples $\mathcal{D}_{O}$.    

The aim of PLL is to learn a classifier $f(\cdot)$ from these partially-labeled examples. Since the inherent label ambiguity in ${Y}$ would impair the training of the classifier, \emph{label disambiguation} is indispensable, which aims to identify the unknown true label $y$ from the candidate label set ${Y}$. Formally, an example $x_{i}$ is associated with a label confidence vector $\boldsymbol{p}_{i} \in [0,1]^{c}$ where each entry denotes the probability of the corresponding label being the ground-truth. The update of $\boldsymbol{p}_{i}$ implies the process of label disambiguation. Ideally, when the $j$-th candidate label of the entry $\boldsymbol{p}_{ij}=1$ is the true label, this indicates the success of label disambiguation. 

The key challenge of this new PLL problem is that the OOC example is inherent to resist label disambiguation, as its true label is outside the candidate label set, which means that label disambiguation for the OOC example fails inevitably, i.e., the update of its $\boldsymbol{p}$ is always misguided. In this case, the OOC example presents a huge threat to the generalization. So, it is non-trivial to handle the particularly harmful OOC example in PLL. Next, we will introduce our method to solve this problem.

\begin{figure*}[!t]
\centering
\includegraphics[width=0.9\textwidth]{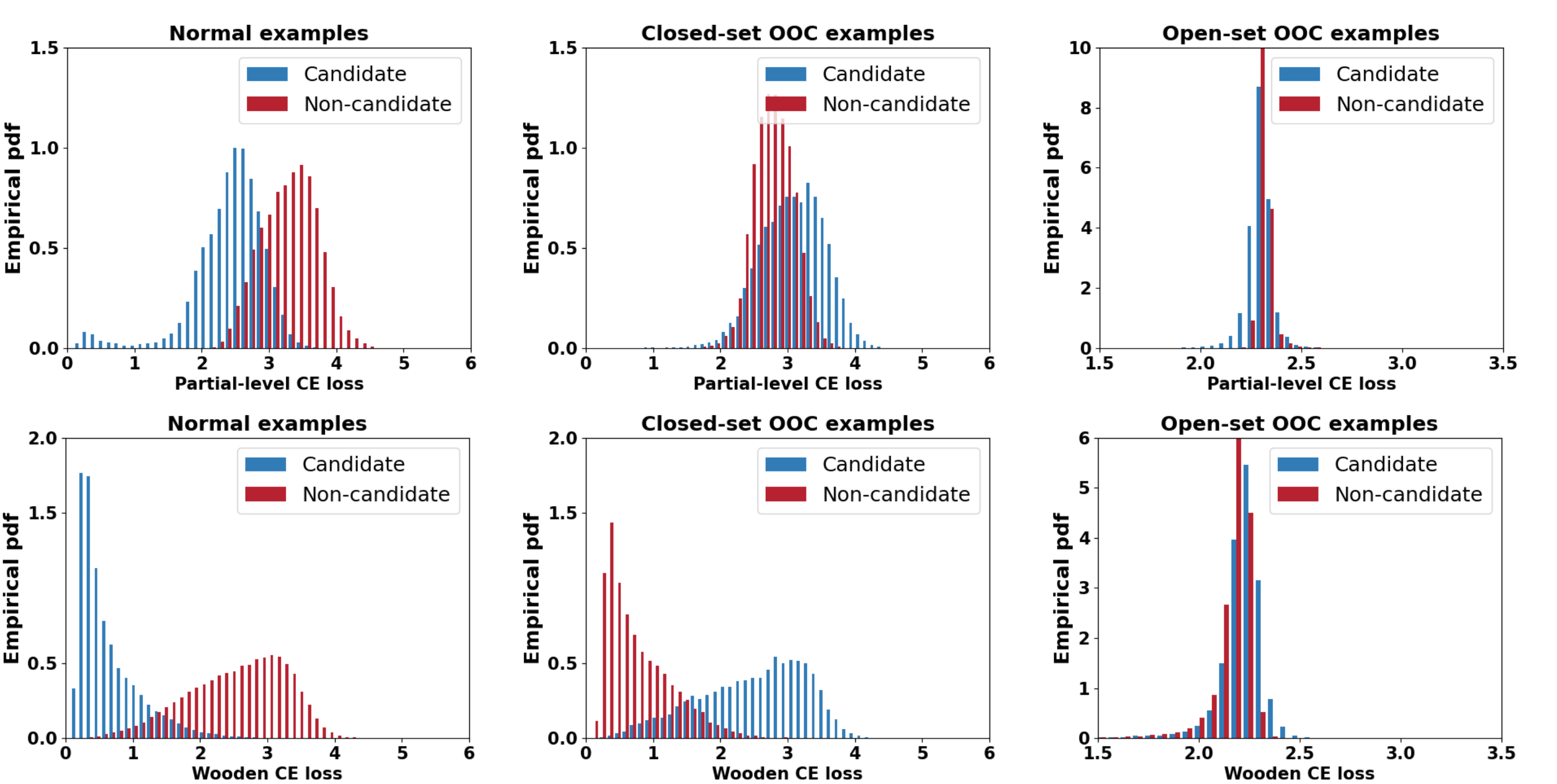} 
	\caption{Partial-level and wooden CE Loss distributions of three types of examples (at epoch 30, $q=0.3, \tau_{1}=0.2, \tau_{2}=0.4$). The blue (red) color denotes the candidate (non-candidate) part.}
	\label{loss_distribution}
\end{figure*}
\begin{figure}[!t]
\centering
\includegraphics[width=0.45\textwidth]{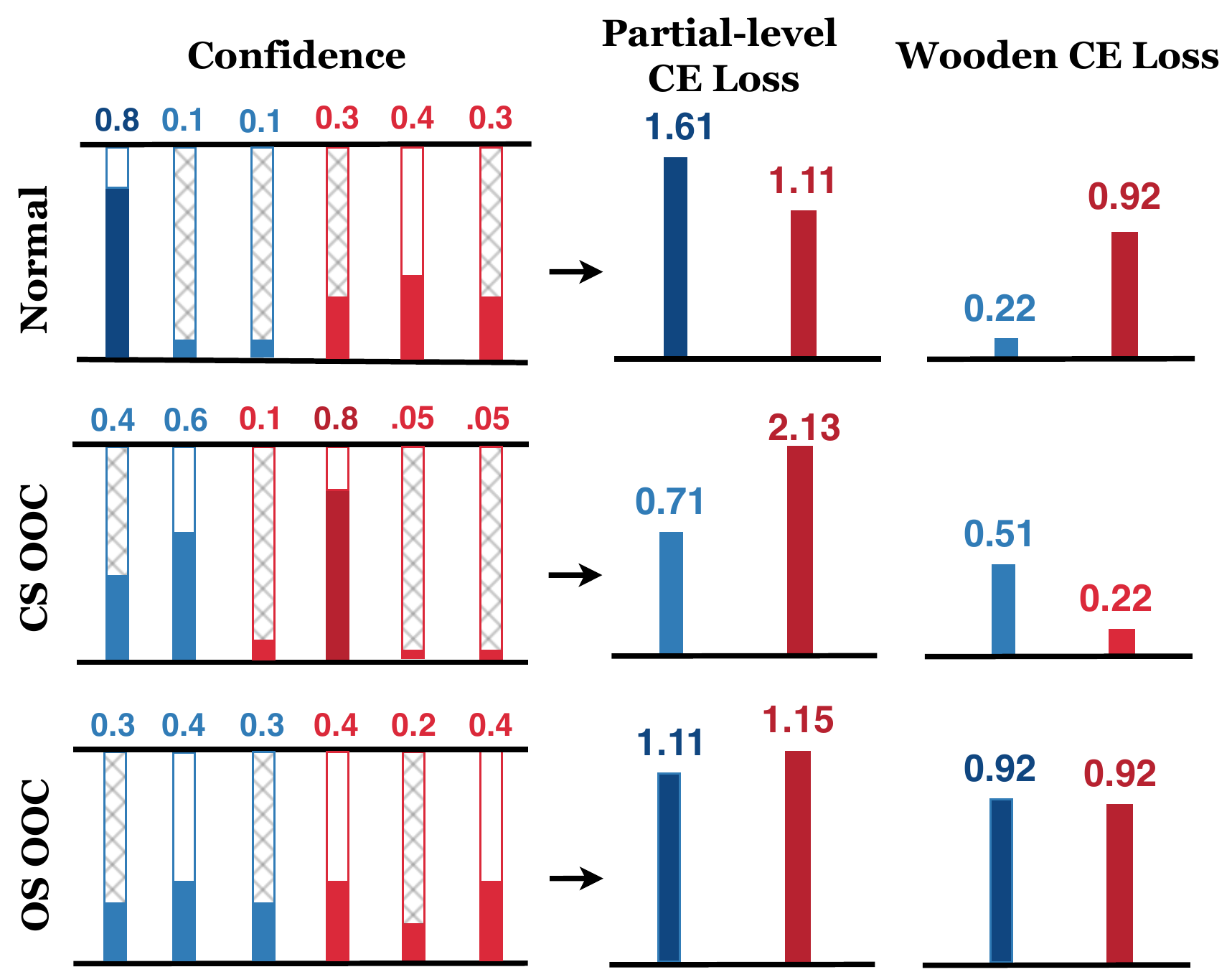} 
	\caption{An example of three types of examples with candidate and non-candidate (wooden) CE loss. The blue (red) color denotes the candidate (non-candidate) part. The trip with a cross-hatch means that we do not use this part to calculate the wooden CE loss.}
	\label{wce}
\end{figure}

\section{Methodology}
In this section, we introduce our method to learn from mixed closed-set and open-set OOC examples. As shown in Figure \ref{framework}, our method mainly includes three parts: OOC selection, (reversed) label disambiguation, and random candidate generation. As mentioned before, OOC examples are greatly harmful in PLL, and thus it is necessary to differentiate them effectively, i.e., termed OOC selection.

\subsection{OOC Selection}
The essence of OOC selection is to determine whether the true label of an example is inside the candidate label set (normal example), the non-candidate label set (closed-set example), or neither (open-set example). For this purpose, small-loss criterion \cite{han2018co}, based on the gradually learning (memorizing) pattern of over-parameterized deep neural networks (DNNs) \cite{zhang2017understanding} has been shown to be effective in (open-set) label-noise learning. But, the difficulty of directly employing this criterion for OOC selection stems from the intrinsic characteristic of PLL, i.e., each partially-labeled example possesses multiple candidate labels instead of only one single label. The core is the calculation of loss for the small-loss criterion. We will show the barrier later and provide our solution for this. To better discriminate the three types of partially-labeled examples, we further decouple the loss into two individual parts: candidate loss $\mathcal{L}$ and non-candidate loss $\widebar{\mathcal{L}}$. The decoupled candidate loss means that its calculation only involves the outputs from candidate labels. Especially, for the decoupled non-candidate loss, we assume that the non-candidate labels are positive, and thus calculate the loss only from non-candidate items. In this way, the candidate (non-candidate) loss has an implication of the probability of a true label inside the candidate (non-candidate) label set (i.e., the smaller the candidate or non-candidate loss, the higher the probability of a true label inside the candidate or non-candidate label set). Based on this, a normal example tends to achieve a smaller candidate loss than a non-candidate one; On the contrary, a closed-set OOC example could possess a smaller non-candidate loss instead; Especially, an open-set example could have both large candidate and non-candidate losses. Therefore, we propose the following OOC selection criterion:   
\begin{equation}
  	\label{selection criterion}
  	\left\{
  		\begin{array}{lll}
    
        \widetilde{\mathcal{D}}_{N}=\mathrm{argmax} _{\mathcal{D}’:|\mathcal{D}’|\geq(1-\gamma _{1})|\mathcal{D}|}(\widebar{\mathcal{L}}(\mathcal{D}’)-\mathcal{L}(\mathcal{D}’)), & \\

        \widetilde{\mathcal{D}}_{C}=\mathrm{argmax} _{\mathcal{D}’:|\mathcal{D}’|\geq(1-\gamma _{1})|\mathcal{D}|}(\mathcal{L}(\mathcal{D}’)-\widebar{\mathcal{L}}(\mathcal{D}’)),  & \\
        
        \widetilde{\mathcal{D}}_{O}=\mathrm{argmax} _{\mathcal{D}’:|\mathcal{D}’|\geq(1-\gamma _{1})|\mathcal{D}|}(\widebar{\mathcal{L}}(\mathcal{D}’)+\mathcal{L}(\mathcal{D}’)),  & 
  		    
  	\end{array}
    \right.
\end{equation} 
where $\gamma_{1}$ ($\gamma_{2}$) is the selection proportion for closed-set (open-set) OOC examples. Then, one straightforward way is using the cross-entropy (CE) loss to instantiate Eq.(\ref{selection criterion}). Formally, given a partially-labeled example $x$ (Note that we have omitted the subscript for the simplification) that has a candidate label set $Y$ with $k$ candidate labels and a non-candidate label set $\widebar{Y}$ with $z$ non-candidate labels ($k+z=c$), the decoupled candidate and non-candidate CE loss are defined as: 
\begin{equation}
    \label{ce}
    \left \{
    \begin{array}{ll}
      l_{c}=\sum_{i=1}^{k}l_{ce}^{i}(f(x),Y)/|Y|,  &   \\
      \widebar{l}_{c}=\sum_{j=1}^{z}l_{ce}^{j}(f(x),\widebar{Y}))/|\widebar{Y}|,    &
    \end{array}
    \right.
\end{equation}
where $l_{ce}^{i}(f(x), Y)=-y^{Y^{i}}log(f^{Y^{i}}(x))$ is the binary item of the CE loss for a single label $Y^{i}$ ( which is the $i$-th candidate label in the candidate label set $Y$ of the example $x$), $y^{m} (m=Y^{i})$ is the $m$-th entry of the label vector $\mathbf{y}$ and $f^{n}(x) (n=Y^{i})$ is the $n$-th item of the network output $f(x)$. Note that the conventional CE loss involves all $c$ items, while the decoupled CE loss only relates to a part of the items. We call it the partial-level CE loss and use it to calculate for the comparison shown in figure \ref{loss_distribution} and \ref{wce}. This formulation implies the average binary CE loss from all candidates (non-candidates). However, we discover that using Eq.(\ref{ce}) for OOC selection is infeasible. This stems from its bias against the sharpness of confidence and label scale. To show this, we provide an illustration shown in Figure \ref{wce}. First, for a normal example (the first line), its candidate CE loss with a \emph{sharp} candidate confidence (the blue part) is larger than its non-candidate loss with a \emph{flat} non-candidate confidence (the red part); for closed-set example (the second line), its non-candidate CE loss a \emph{sharp} candidate confidence (the red part) is also larger than its candidate counterpart (the blue part). This is inconsistent with the fact (i.e., the sharper the confidence, the smaller the loss), which implies the bias against the sharpness of confidence. Second, compared the normal example's candidate part (the blue in the first line) with the closed-set example's non-candidate part (the red part in the second line), with the nearly same sharpness, the loss in the latter with one more label is larger than that in the former dramatically. This reveals the bias against label scale, i.e., the magnitude of CE loss is proportional to the number of labels (under the nearly same sharpness of confidence). Consequently, the simple decoupled CE loss is not qualified for OOC selection. 

To solve this issue, we utilize the \textit{wooden} CE loss motivated by the "wooden buckets effect":
\begin{small}
\begin{equation}
    \label{wce_loss}
    \left \{
    \begin{array}{ll}
    l_{w}=\mathrm{min}(l_{ce}^{1}(f(x),Y),\cdots,l_{ce}^{k}(f(x),Y)), & \\
    \widebar{l}_{w}=\mathrm{min}(l_{ce}^{1}(f(x),\widebar{Y}),\cdots,l_{ce}^{z}(f(x),\widebar{Y})). & 
       \end{array}
    \right.
\end{equation}
\end{small}This implies that the candidate or non-candidate loss is only determined only by the \emph{minimum} item of the decoupled CE loss from candidates or non-candidates. Surprisingly, it has two good properties against the aforementioned bias. First, from the wooden CE loss of normal and closed-set OOC examples shown in Figure \ref{wce}, we can see that it is consistent with the fact (i.e., the sharper the confidence, the smaller the loss). Second, comparing the wooden candidate CE loss of normal example with the wooden non-candidate CE loss of closed-set OOC example, we can see that it is invariant to the label scale. Intrinsically, a smaller wooden CE loss in candidate or non-candidate implies that label disambiguation is favorable in the candidate or non-candidate label set. (i.e., with a \emph{sharp} disambiguated label distribution) and thus true label is most likely inside the candidate or non-candidate label set. 

Particularly, a potential alternative solution is to utilize the entropy of confidence from the candidate or non-candidate items. That is to say the lower the entropy of confidence from candidates (non-candidates), the higher probability of the true label inside the candidate label set (the non-candidate label set). But, the thorny issue lies in that the number of an example's candidate labels greatly influences the magnitude of its corresponding entropy. Therefore, it is considerably difficult to discriminate them by a standard entropy threshold value. Otherwise, we consider the wooden loss to achieve the aim.

\noindent\textbf{Moving-average ensemble output.} Furthermore, to improve the stability and precision of OOC selection, we use the moving-average ensemble output to calculate the wooden CE loss. First, the ensemble output of an example $x$ is obtained by averaging the over $\phi$ training epochs: $\widebar{f}(x)=\sum_{i=1}^{\phi}f^{(i)}(x)/\phi$ where $f^{(i)}(x)$ denotes the model output at epoch $i$. After this, we update the ensemble output by a moving-averaging means: $\widebar{f}^{(i)}(x)=\eta\widebar{f}^{(i-1)}(x)+(1-\eta)f^{(i)}(x)$, whereby $\eta$ is a momentum. In this way, we utilize the wooden CE loss Eq.(\ref{wce_loss}) to instantiate Eq.(\ref{selection criterion}). To verify our intuition of OOC selection, we perform a quantitative experiment to show the difference between ordinary CE loss distribution and wooden CE loss distribution shown in Figure \ref{loss_distribution}. We can observe that it is indistinguishable for normal and closed-set OOC examples based on ordinary partial-level CE loss, while that of wooden CE loss is well-distinguished. This definitely justifies our motivation.    

Now, after selecting these OOC examples $\widetilde{\mathcal{D}}_{C}$ and $\widetilde{\mathcal{D}}_{O}$, the next challenge is how to handle them effectively. An intuitive subsequent means is to discard them to prevent them from harming the training of DNNs. Such a way of combating OOC examples is straightforward but obviously sub-optimal since these abandoned OOC examples also contain useful information for generalization. Therefore, we leverage them for training in different ways. Next, we introduce reversed label disambiguation for close-set OOC examples.  
\subsection{Reversed Label Disambiguation}
Label disambiguation is an indispensable way of combating label ambiguity in PLL. Ordinary label disambiguation for $\widetilde{\mathcal{D}}_{N}$ is shown as: 
\begin{equation}
	\label{ld}
	p_{ij}=\left\{
	\begin{array}{cl}
		\frac{f_{j}(x_{i})}{\sum_{j=1}^{c}f_{j}(x_{i})}, & if \quad j \in Y_{i},  \\
		0, &if \quad j \in \widebar{Y}_{i}, 
	\end{array}
	\right.
\end{equation}
where $f_{j}(x_{i})$ is the $j$-th coordinate of the network output $f(x_{i})$. The first term reveals the nature of label disambiguation that pushes more weights to more possible candidate labels slightly and progressively. The second term means that we do not consider non-candidate labels, and thus their labeling confidence maintains zero. Based on this equipment, we can calculate the loss against $\widetilde{\mathcal{D}}_{N}$: 
\begin{equation}
    \label{loss_normal}
    \mathcal{L}_{N}=-\frac{1}{|\mathcal{B}|}\sum\limits_{(x_{i},\mathbf{p}_{i})\in\widetilde{\mathcal{D}}_{N}}\sum_{j=1}^{c}p_{ij}\mathrm{log}f_{j}(x_{i}).
\end{equation}
Where $|\mathcal{B}|$ is the size of a batch. However, Eq. (\ref{ld}) is inapplicable for selected close-set OOC examples $\widetilde{\mathcal{D}}_{C}$, since their true label is outside their candidate label set. To solve this issue, we propose \textit{reversed label disambiguation} that identifies the true label from the non-candidate label set instead for $\widetilde{\mathcal{D}}_{C}$: 
\begin{equation}
	\label{rld}
	\widebar{p}_{ij}=\left\{
	\begin{array}{cl}
		\frac{f_{j}(x_{i})}{\sum_{j=1}^{c}f_{j}(x_{i})}, & if \quad j \in \widebar{Y}_{i},  \\
		0, &if \quad j \in Y_{i}. 
	\end{array}
	\right.
\end{equation}
The only change of Eq. (\ref{rld}) is the switched condition, which aims to perform label disambiguation in the non-candidate label set instead. Now, we can calculate the loss against $\widetilde{\mathcal{D}}_{C}$: 
\begin{equation}
    \label{loss_cs}
    \mathcal{L}_{C}=-\frac{1}{|\mathcal{B}|}\sum\limits_{(x_{i},\widebar{\mathbf{p}}_{i})\in\widetilde{\mathcal{D}}_{C}}\sum_{j=1}^{c}\widebar{p}_{ij}\mathrm{log}f_{j}(x_{i}).
\end{equation}
Now, the only remaining question is how to handle open-set OOC examples. Both label disambiguation (i.e., Eq.(\ref{ld})) and reversed label disambiguation (i.e., Eq.(\ref{rld})) are invalid for open-set OOC examples, as their true label is outside the defined classes. Next, we show our solution to further leverage them for training.

\begin{algorithm}[t]
	\caption{Our Method}
	\label{algorithm}
	\LinesNumbered 
	\KwIn{Partially-labeled dataset ${\mathcal{D}}$, model $f(\cdot)$, parameters: trade-off parameters $\alpha$ and $\beta$, selection proportion $\gamma_{1}$, $\gamma_{2}$, the epoch of warm-up: $T_{warmup}$.} 
	\KwOut{model parameters}
	\For{$t<T_{max}$}{
	    \For{$t<T_{warmup}$}{
	        Warm up the model for a period of epochs;\\
	        Record the model output at each epoch;\\
	    }
	    Obtain the ensemble output;\\
		\For{$t>=T_{warmup}$}{
		    Update the moving-averaging ensemble output;\\
		    Use Eq.(1) to divide the dataset into $\widetilde{\mathcal{D}}_{N}$,  $\widetilde{\mathcal{D}}_{C}$ and  $\widetilde{\mathcal{D}}_{O}$;\\
            \For{$b=1\ \mathrm{to}\ \mathcal{B}$}{
                Ordinal label disambiguation by Eq.(4) for $\widetilde{\mathcal{D}}_{N}$;\\
                Reversed label disambiguation by Eq.(6) for $ \widetilde{\mathcal{D}}_{C}$;\\
                Random candidate generation for $ \widetilde{\mathcal{D}}_{O}$;\\
                Calculate the overall loss by Eq.(9);\\
            }
		}
	}
\end{algorithm}

\begin{table*}[!t]
    \small
	\centering
	\caption{Test accuracy comparisons. Bold indicates superior results. “C+S” is the abbreviation of CIFAR-10+SVHN. “C+C” is the abbreviation of CIFAR-10+CIFAR-100. “C+I” is the abbreviation of CIFAR-10+ImageNet32.}
	\begin{tabular}{c|cr|cc|cc|cc}
		\toprule
		\multirow{2}{*}{Dataset} 
		&\multirow{2}{*}{Method} &$q$           & \multicolumn{2}{c}{0.1}          &\multicolumn{2}{c}{0.3}                &\multicolumn{2}{c}{0.5}  \\ 
		&                        &$\tau$        & 0.2$\And$0.4  &0.3$\And$0.6     & 0.2$\And$0.4  &0.3$\And$0.6       & 0.2$\And$0.4 &0.3$\And$0.6              \\ \midrule
		\multirow{5}{*}{C+S}   
		& CC\cite{feng2020provably}    &   &82.84 $\pm$ .09 &74.77 $\pm$ .12   &78.47 $\pm$ .20 &68.37 $\pm$ .15    &67.65 $\pm$ .25 &47.13 $\pm$ .30 \\
		& PRODEN\cite{feng2020provably}&   &83.28 $\pm$ .14 &75.43 $\pm$ .05   &78.63 $\pm$ .15 &69.06 $\pm$ .25    &71.53 $\pm$ .13 &52.14 $\pm$ .15 \\
		& LWS\cite{wen2021leveraged}   &   &80.33 $\pm$ .12 &74.96 $\pm$ .10   &77.47 $\pm$ .13 &69.07 $\pm$ .30    &71.84 $\pm$ .11 &53.79 $\pm$ .12 \\
            & CAVL \cite{zhang2021exploiting} &  &76.82 $\pm$ .12 & 68.72 $\pm$ .19 & 66.28 $\pm$ .35 & 56.32 $\pm$ .28  & 60.12 $\pm$ .26 & 43.66 $\pm$ .15 \\
		& PiCO \cite{wang2022pico}     &   &84.15 $\pm$ .20 &76.22 $\pm$ .08   &81.40 $\pm$ .09 &72.21 $\pm$ .11    &77.01 $\pm$ .10 &61.88 $\pm$ .25 \\
            & CRDPLL \cite{wu2022revisiting}   &   &85.62 $\pm$ .25 &81.26 $\pm$ .31   &83.13 $\pm$ .18 &76.84 $\pm$ .17    &78.09 $\pm$ .28 & 63.28 $\pm$ .41 \\
		&  \cellcolor{gray!25} Ours    &\cellcolor{gray!25}   &\cellcolor{gray!25}  \textbf{91.30 $\pm$ .10} &\cellcolor{gray!25} \textbf{90.20 $\pm$ .25 }  &\cellcolor{gray!25} \textbf{87.85 $\pm$ .12 }&\cellcolor{gray!25} \textbf{83.28 $\pm$ .22}    &\cellcolor{gray!25} \textbf{79.54 $\pm$ .36} & \cellcolor{gray!25}\textbf{64.10 $\pm$ .14} \\  \midrule
		\multirow{2}{*}{Dataset} 
		&\multirow{2}{*}{Method} &$q$           & \multicolumn{2}{c}{0.1}          &\multicolumn{2}{c}{0.3}                &\multicolumn{2}{c}{0.5}  \\ 
		&                        &$\tau$        & 0.2$\And$0.4  &0.3$\And$0.6     & 0.2$\And$0.4  &0.3$\And$0.6       & 0.2$\And$0.4 &0.3$\And$0.6              \\ \midrule
		\multirow{5}{*}{C+C}   
		& CC\cite{feng2020provably}    &   &80.36 $\pm$ .25 &70.94 $\pm$ .18   &76.20 $\pm$ .35 &64.74 $\pm$ .14    &62.61 $\pm$ .06 &40.18 $\pm$ .15 \\  
		& PRODEN\cite{feng2020provably}&   &80.46 $\pm$ .08 &71.25 $\pm$ .22   &76.24 $\pm$ .15 &65.91 $\pm$ .20    &68.28 $\pm$ .11 &48.77 $\pm$ .24 \\
		& LWS\cite{wen2021leveraged}   &   &80.79 $\pm$ .10 &71.67 $\pm$ .05   &77.31 $\pm$ .72 &65.38 $\pm$ .25    &67.83 $\pm$ .20 &48.60 $\pm$ .10 \\
            & CAVL \cite{zhang2021exploiting} & &74.68 $\pm$ .10 & 67.48 $\pm$ .15 & 64.32 $\pm$ .23 & 54.46 $\pm$ .17  & 58.37 $\pm$ .15 & 42.58 $\pm$ .29 \\
		& PiCO \cite{wang2022pico}     &   &83.63 $\pm$ .14 &76.36 $\pm$ .52   &81.10 $\pm$ .27 &72.33 $\pm$ .08    &75.42 $\pm$ .18 &60.35 $\pm$ .15 \\
            & CRDPLL \cite{wu2022revisiting} & & 84.21 $\pm$ .10 & 79.32 $\pm$ .15 & 82.36 $\pm$ .20 & 75.24 $\pm$ .15  & 75.20 $\pm$ .12 & 60.25 $\pm$ .12 \\
		&  \cellcolor{gray!25} Ours     &\cellcolor{gray!25}   &\cellcolor{gray!25}  \textbf{90.05 $\pm$ .18} &\cellcolor{gray!25}\textbf{87.90 $\pm$ .30 }  &\cellcolor{gray!25}\textbf{85.25 $\pm$ .07 }&\cellcolor{gray!25}\textbf{79.75 $\pm$ .30}    &\cellcolor{gray!25}\textbf{76.56 $\pm$ .22} &\cellcolor{gray!25}\textbf{60.88 $\pm$ .25} \\  \midrule
		\multirow{2}{*}{Dataset} 
		&\multirow{2}{*}{Method} &$q$           & \multicolumn{2}{c}{0.1}          &\multicolumn{2}{c}{0.3}                &\multicolumn{2}{c}{0.5}  \\ 
		&                        &$\tau$        & 0.2$\And$0.4  &0.3$\And$0.6     & 0.2$\And$0.4  &0.3$\And$0.6       & 0.2$\And$0.4 &0.3$\And$0.6              \\ \midrule
		\multirow{5}{*}{C+I}   
	    & CC\cite{feng2020provably}    &   &83.22 $\pm$ .15 &75.45 $\pm$ .10   &78.71 $\pm$ .12 &67.75 $\pm$ .21    &66.42 $\pm$ .52 &46.53 $\pm$ .15 \\
		& PRODEN\cite{feng2020provably}&   &83.68 $\pm$ .22 &75.17 $\pm$ .22   &78.96 $\pm$ .09 &68.89 $\pm$ .13    &71.26 $\pm$ .20 &51.79 $\pm$ .10 \\
		& LWS\cite{wen2021leveraged}   &   &84.18 $\pm$ .20 &75.21 $\pm$ .28   &79.90 $\pm$ .23 &69.73 $\pm$ .14    &73.21 $\pm$ .27 &53.42 $\pm$ .23 \\
            & CAVL \cite{zhang2021exploiting} & &76.15 $\pm$ .20 & 67.84 $\pm$ .25 & 66.83 $\pm$ .20 & 55.46 $\pm$ .18  & 60.75 $\pm$ .10 & 43.48 $\pm$ .12 \\
		& PiCO \cite{wang2022pico}     &   &84.65 $\pm$ .16 &76.18 $\pm$ .20   &81.55 $\pm$ .15 &72.65 $\pm$ .12    &77.48 $\pm$ .38 &55.63 $\pm$ .15 \\
            & CRDPLL \cite{wu2022revisiting} &  & 84.46 $\pm$ .13 & 82.44 $\pm$ .21  & 82.70 $\pm$ .23 & 74.92 $\pm$ .15  & 77.52 $\pm$ .31 & 63.28 $\pm$ .41 \\
		&  \cellcolor{gray!25} Ours     &\cellcolor{gray!25}   &\cellcolor{gray!25}  \textbf{91.54 $\pm$ .10} &\cellcolor{gray!25}\textbf{89.83 $\pm$ .25 }  &\cellcolor{gray!25}\textbf{87.68 $\pm$ .15 }&\cellcolor{gray!25}\textbf{83.88 $\pm$ .10}    &\cellcolor{gray!25}\textbf{79.27 $\pm$ .14} &\cellcolor{gray!25}\textbf{64.98 $\pm$ .18} \\  \bottomrule
	\end{tabular}
	\label{result}
\end{table*}

\subsection{Random Candidate Generation}
Generally, in label-noise learning, open-set example always plays a negative role in the training process \cite{wang2018iterative}, which are usually considered to be harmful to the training of DNNs. Several related methods are proposed to reduce the negative effect of open-set noisy examples by filtering out them \cite{wang2018iterative,sachdeva2021evidentialmix} or modeling the extended transition matrix \cite{xia2022extended}. On the contrary, recent work \cite{wei2021open} reveals a counter-intuitive observation that open-set noisy examples can even benefit the robustness of DNNs against noisy labels. Concretely, they propose to generate one random label for each auxiliary open-set example over training epochs. The intuitive interpretation based on the "insufficient capacity" hypothesis is that fitting these random open-set noisy examples consumes extra capacity of DNNs, thereby reducing the memorization of inherent noisy labels. Following this point, we propose random candidate generation that dynamically assigns a random candidate label set $Y^{O}$ for each selected open-set OOC example $x\in\widetilde{\mathcal{D}}_{O}$. In this way, DNNs tend to consume extra capacity to disambiguate them, thus alleviating the memorization of falsely disambiguated examples in (reversed) label disambiguation. The loss against $\widetilde{\mathcal{D}}_{O}$ is calculated by: 
\begin{equation}
    \label{loss_os}
    \mathcal{L}_{O}=\frac{1}{|\mathcal{B}|}\sum\limits_{(x_{i},Y_{i}^{O})\in\widetilde{\mathcal{D}}_{O}}l_{CE}(f(x),Y^{O}),
\end{equation}
where $l_{CE}(f(x),Y)=-\sum_{j}y_{j}log(f_{j}(x))$ is the CE loss for the label set $Y$. Finally, the overall loss is shown as: 
\begin{equation}
    \label{all_loss}
    \mathcal{L}=\mathcal{L}_{N}+\alpha\mathcal{L}_{C}+\beta\mathcal{L}_{O},
\end{equation}
where $\alpha$ and $\beta$ are two trade-off parameters to balance the loss. Note that the overall loss is calculated in a batch-wise training procedure. The pseudo-code of the proposed method is shown in algorithm 1.
\begin{figure*}[!t]
\centering
\includegraphics[width=0.9\textwidth]{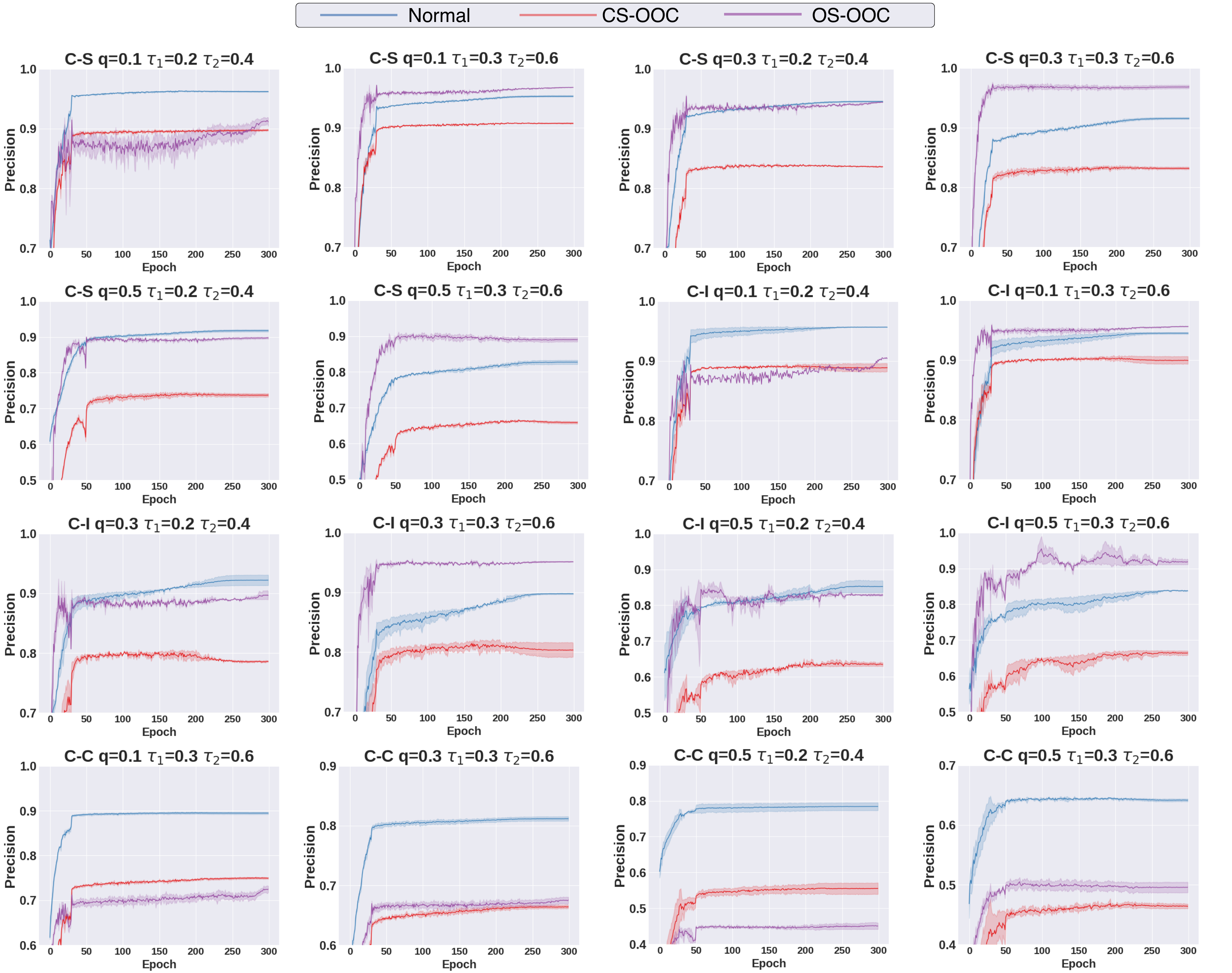} 
	\caption{Precision curves of OOC selection against selected normal (red), closed-set (blue), and open-set (purple) examples on CIFAR-10-SVHN, CIFAR-10-ImageNet32, and CIFAR-10-CIFAR-100.}
	\label{precision}
\end{figure*}

\section{Experiments}
In this section, we conduct extensive experiments with benchmark datasets: CIFAR-10, CIFAR-100, SVHN, and ImageNet32 to demonstrate the superiority and effectiveness of our proposed method. We first introduce the setup. 
\subsection{Setup}
\noindent \textbf{Datasets.}\quad We evaluate our proposed method on synthetic CIFAR-10 with comprehensive experiments. First, similar to previous work \cite{feng2020provably,lv2020progressive}, we generate partially-labeled examples by a uniform label flipping probability $q$ to generate the candidate label set. The value of $q$ is in [0.1, 0.3, 0.5] where larger $q$ denotes a higher label ambiguity degree. For the generation of close-set OOC example, we randomly select $\tau_{1}\%$ examples in CIFAR-10 and replace their true label with a random non-candidate label, which does not change the number of candidate labels. To generate the open-set OOC example, we use auxiliary datasets SVHN, CIFAR-100, and ImageNet32. SVHN is a real-world image dataset for developing machine learning and object recognition algorithms with minimal data preprocessing and formatting requirements. ImageNet32 is a huge dataset made up of small images called the down-sampled version of ImageNet. Concretely, we randomly add $\tau_{2}\%$ examples from auxiliary datasets into the original training data. In this way, the example size becomes $50000\times(1+\tau_{2})$. We consider $\tau_{1}=[0.2, 0.3]$ and $\tau_{2}=[0.4, 0.6]$ respectively.

\noindent \textbf{Compared methods.}\quad Four state-of-the-start deep partial label learning algorithms are compared: (1) CC \cite{feng2020provably} is a classifier-consistent method that assumes set-level uniform data generation process; (2) PRODEN \cite{lv2020progressive} progressively identifies the ground-truth label with a self-training style way; (3) LWS \cite{wen2021leveraged} weights the risk function by means of a trade-off between losses on candidates and non-candidates. We use the suggested parameters in the original paper; (4) CAVL \cite{zhang2021exploiting} exploits the class activation value for label disambiguation; (5) PiCO \cite{wang2022pico} uses contrastive learning technologies to label disambiguation better; (6) CRDPLL \cite{wu2022revisiting} utilizes the technology of consistency regularization and achieves the state-of-the-art performance in partial-label learning. Note that the first four methods (i.e., CC, PRODEN, LWS, and CAVL) do not use the powerful data augmentation technology in the original paper. To achieve a fair comparison with a slight modification, we use the same data augmentation technology for all methods in this paper,  thereby resulting in slight differences in performance.  

\noindent \textbf{Implementation.}\quad For the fair comparison, we implement all methods with the same base training scheme: a ResNet-18 architecture, a standard SGD optimizer with a momentum of 0.9 and weight decay is 0.001, and the learning rate is 0.01 with a cosine learning rate schedule, a total training epoch 300. The batch size is 128. The momentum parameter $\eta$ is set to 0.9. The parameters $\alpha$ and $\beta$ are selected from [0.1, 0.5, 1, 3, 5] and set to 1, 0.1 respectively. The epoch of warm-up is set to 30 (50 for $q=0.5$). The ensemble epoch $\phi$ is set to 5 (20 for $q=0.5$). We set the parameters $\gamma_{1}$ and $\gamma_{2}$ to the value of $\tau_{1}$ and $\tau_{2}$ respectively. Specially, we use the same data augmentation technology for all methods. We present the mean and standard deviation in each case based on five trials. 

\begin{table}[t]
    \small
	\centering	
	\caption{Ablation study on CIFAR-10-SVHN ($q=0.1$).}
	\begin{tabular}{c|cccc}
		\toprule
		\multirow{2}{*}{Ablation}     &\multicolumn{2}{c}{Label Disambiguation} & \multicolumn{2}{c}{($\tau_{1} \tau_{2}$)} \\
		 &CS-OOC       &OS-OOC          & (0.2 0.4)            & (0.3 0.6)           \\ \midrule
		 w/o LD       &$\times$   &$\times$  &72.92 $\pm$ .15     &63.05 $\pm$ .18    \\
		 w/o WCE       &RLD        &RCG       &89.68 $\pm$ .23     &89.15 $\pm$ .12    \\
		 w/o Warm up   &RLD        &RCG       &72.36 $\pm$ .18     &64.85 $\pm$ .21   \\
		 w/o RCG       &RLD        &LD        &90.47 $\pm$ .15     &89.32 $\pm$ .17   \\   
		 w/o RLD       &LD         &RCG       &83.46 $\pm$ .11     &76.13 $\pm$ .06   \\  \midrule
		 Ours          &RLD        &RCG       &  \textbf{91.52 $\pm$ .08}        &\textbf{90.21 $\pm$ .12 }        \\ \bottomrule
	\end{tabular}
	\label{as}
\end{table}

\subsection{Experimental Results}

\noindent \textbf{Accuracy comparison.}\quad As shown in Table \ref{result}, our method outperforms all counterparts under all cases with a significant gap. For each fixed partial rate (i.e., 0.1, 0.3, and 0.5), with the increase of $\tau_{1}$ and $\tau_{2}$ (i.e., more and more OOC examples), all the methods show a significant performance drop (especially for the high partial rate i.e., $q=0.5$). For example, CRDPLL drops by about 2\%, 8\%, and 14\% ($q=0.1, 0.3, 0.5$ on CIFAR-10-ImageNet32) respectively; PiCO drops by about 8\%, 9\%, and 22\% ($q=0.1, 0.3, 0.5$ on CIFAR-10-ImageNet32) respectively. This phenomenon implies OOC examples show a great threat against the ordinal partial-label learning methods even for the state-of-the-art ones. Meanwhile, this observation definitely verifies the significance of our claim to deal with OOC examples. On the other hand, for a fixed proportion ($\tau_1$ and $\tau_2$) of OOC examples, as the partial rate $q$ increases, the performance also drops consistently. This is because the difficulty of label disambiguation increases with the larger partial rate so that the negative effect of OOC examples is amplified. Especially, for both of these two cases above, our method keeps the superiority consistently. This result owes to our strategy against OOC examples which is lacking in compared methods. Furthermore, for different auxiliary open-set datasets, the performance with CIFAR-100 (i.e., C-C) shows a more significant drop compared with C-S, and C-I. This is because images from CIFAR-100 have similar visual information against CIFAR-10, and thus it is easy for the model to overfit them. This observation also throws a concern about the visual-similar OOC example for the model training.  

\noindent \textbf{Evaluation of OOC selection.}\quad As shown in Figure \ref{precision}, we report the precision rate of OOC selection against normal, closed-set, and open-set examples respectively on CIFAR-10-SVHN, CIFAR-10-ImageNet32 and CIFAR-10-CIFAR-100. Note that in our case, precision and recall rate are equivalent, since OOC selection is fixed in proportion. We can observe that the selection performance maintains a high level (e.g., around 95\%, 90\%, 96\% for normal, CS-OOC, OS-OOC respectively on $q=0.1, \tau_{1}=0.3, \tau_{2}=0.6$). Especially, as $q$ increases, the OOC selection performance of normal and open-set examples maintains a stable level, while that of closed-set ones drops sharply. This reveals that high label ambiguity affects the selection of closed-set examples more than others. For CIFAR-10-CIFAR-100, we can observe that the selection performance of normal examples (blue) maintains a high level always. As $q$ increases, the OOC selection performance of closed-set and open-set examples drops sharply, while the performance drops overall against open-set OOC examples. This is because the images in CIFAR-100 are visually similar to that in CIFAR-10, thus improving the difficulty of discriminating them.

\subsection{Ablation Studies}
In this section, we present our ablation studies to show the effectiveness of our method. As shown in Table \ref{as}, the term ``W/o LD (Label Disambiguation)" means that we do not disambiguate examples anymore, which makes the performance drop dramatically and thus shows the significance of LD in PLL; the term ``W/o WCE (wooden cross-entropy)" means that we perform OOC selection with ordinary cross-entropy loss (Eq.(\ref{ce})) instead of the wooden cross-entropy loss (Eq.(\ref{wce})). This observation stems from that OOC selection based on ordinary cross-entropy loss would falsely select many OOC examples as normal ones. In this case, falsely selected OOC examples have a great negative effect on the model training, leading to a significant drop in performance. Therefore, the wooden cross-entropy loss is a better metric than the ordinary cross-entropy loss to discriminate normal and OOC examples. The term "W/o Warm up" means that we do not warm up the network with LD, which also leads to a decline in performance. This observation implies sufficient warming up is important for OOC selection and the whole model training. The term ``w/o RCG" (Random Candidate Generation) means that we do not generate random candidate labels and only use originally assigned candidate labels for selected open-set OOC examples, which affects the original representation learning in CIFAR-10; the term "w/o RLD" (Reversed Label Disambiguation) means that we do not use Eq. (\ref{rld}) but employ Eq. (\ref{ld}) to update selected closed-set OOC examples, which drops the performance sharply, and definitely verifies the superiority of RLD against LD. Therefore, the effectiveness of the proposed method is verified.

\begin{figure}[t]
\centering
\includegraphics[width=0.45\textwidth]{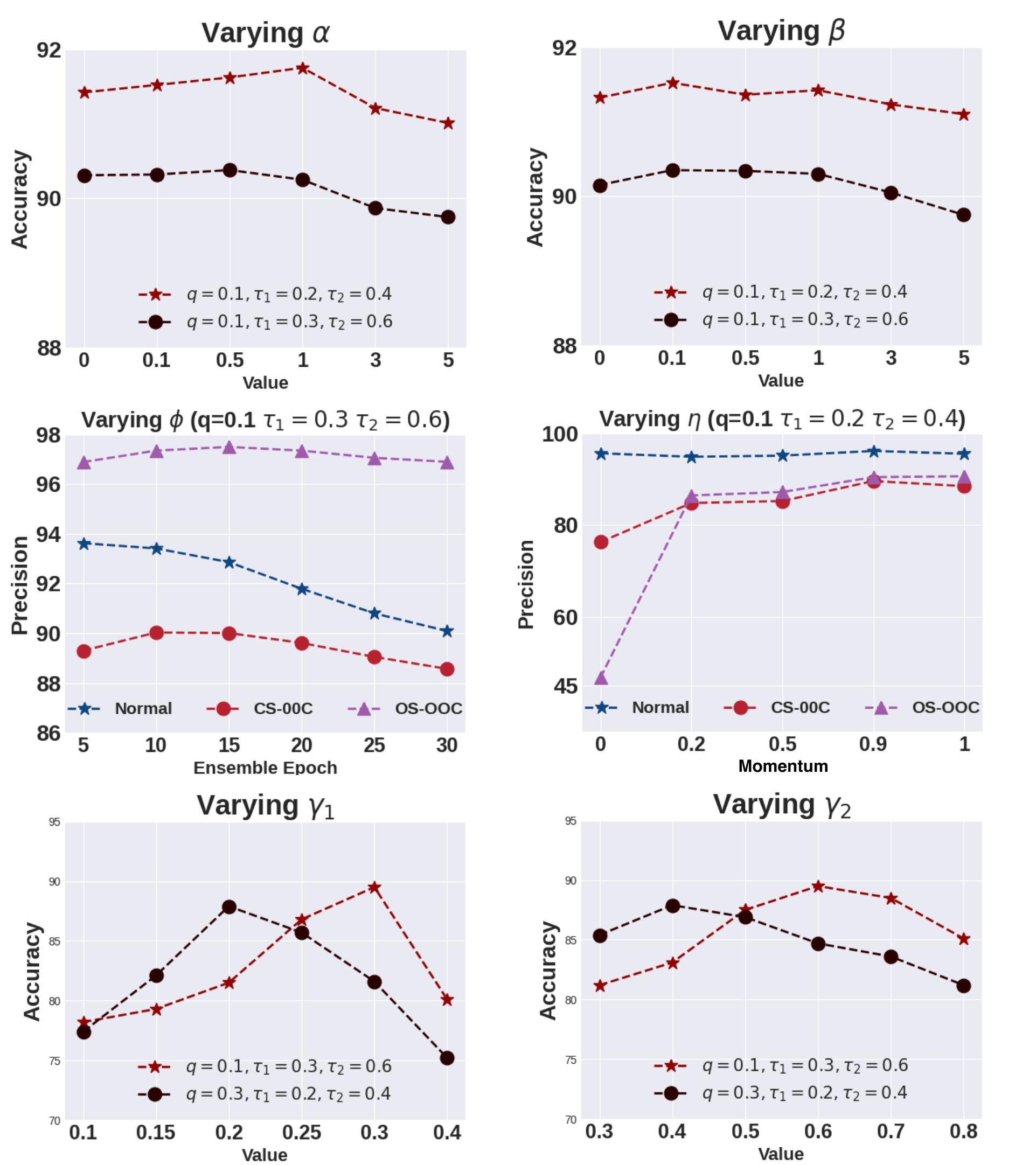} 
	\caption{Varying hyper-parameters ($\alpha$, $\beta$, $\phi$, $\eta$ and $\gamma$).}
	\label{eval_params}
\end{figure}

\begin{figure}[!t]
\centering
\includegraphics[width=0.45\textwidth]{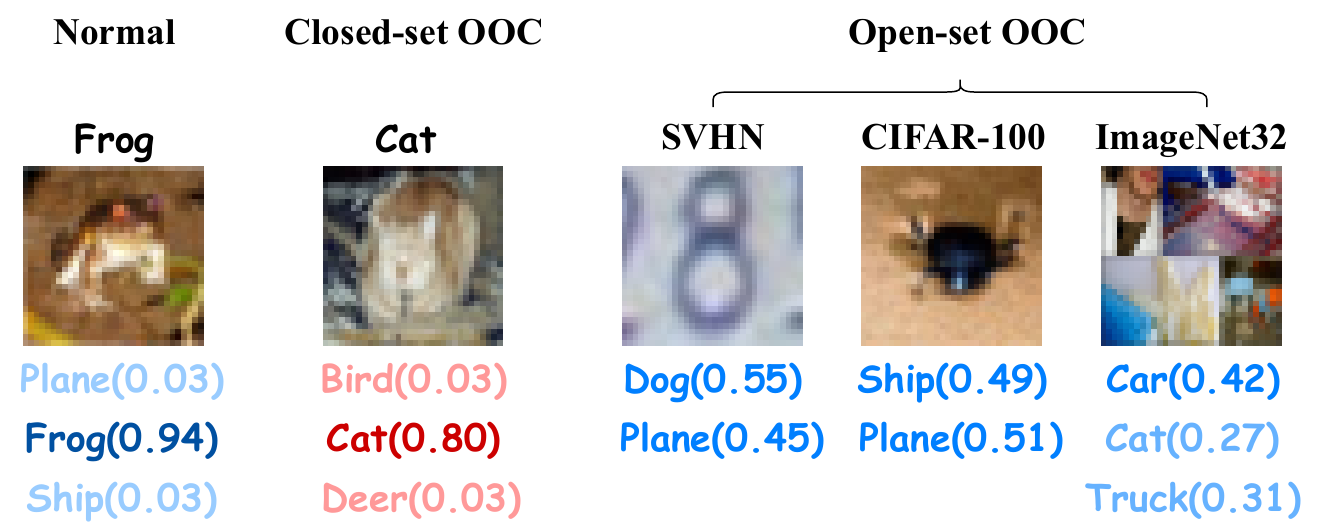} 
	\caption{ Selected disambiguated examples. The blue color denotes the candidate labels, while the red color denotes the non-candidate labels. The darker the color, the higher the confidence.}
	\label{example_img}
\end{figure}
\subsection{Hyper-parameter Analysis}
Here, we mainly evaluate two trade-off parameters $\alpha$, $\beta$, the ensemble epoch $\phi$, the momentum value $\eta$ and the selection proportion $\gamma$. From Figure \ref{eval_params}, for the parameter $\alpha$, we can observe that when $\alpha$ is too large, the negative effect of falsely selected closed-set OOC examples would be amplified greatly, thereby leading to performance degradation. Similarly, the parameter $\beta$ shows a roughly analogical property. Therefore, it is of vital importance to choose the proper values for the parameters $\alpha$ and $\beta$. For the ensemble epoch $\phi$, as $\phi$ increases, the ensemble output contains more and more information from epochs. However, it is not always beneficial for OOC selection. When $\phi$ is too large, the ensemble output contains too many inaccurate outputs from early epochs, which drops the performance of OOC selection dramatically. Otherwise, the ensemble output is not distinguishable enough for OOC selection, which also leads to a drop in precision. For the momentum value $\eta$, when $\eta=0$, we do not use the ensemble output and only employ the output at each epoch for OOC selection, which drops the performance dramatically. This definitely verifies the significance of ensemble output. As $\eta$ increases, the precision improves continually. Especially, when $\eta=1$, it means that we only use the ensemble output alone, which also drops the precision slightly. For the selection proportion $\gamma_{1}$ and $\gamma_{2}$, we directly set these two parameters to the known proportion values $\tau_{1}$ and $\tau_{2}$ in our proposed method. Here, we evaluate these two parameters as shown in Figure \ref{eval_params}. We can see that small selection proportion values lead to degraded performance because too small OOC examples are selected and handled. On the other hand, large selection proportion values also lead to inferior performance. This is because a part of normal examples is falsely selected as OOC ones. Hence, this validates the effectiveness of a moving-average ensemble. 

\subsection{Visualization}
As shown in Figure \ref{example_img}, there are three types of selected disambiguated examples at the end of training. The normal example (its true label is \emph{Frog}) has an assigned candidate label set: \{ \emph{Plane}, \emph{Frog}, \emph{Ship} \}. Our method successfully identified its true label with high label confidence (0.94). The closed-set OOC examples (its true label is \emph{Cat}) has a non-candidate label set: \{ \emph{Bird}, \emph{Cat}, \emph{Deer} \}. Especially, its true label is outside its candidate label set. So, it inevitability fails to identify the true label by ordinary label disambiguation. Luckily, in our method, it is achieved by reversed label disambiguation. That is to say that we identify the true label \emph{Cat} in a non-candidate label set with high label confidence (0.80). In this way, this closed-set OOC example is able to be further leveraged for model training. For the open-set OOC example, they are from three different datasets: SVHN, CIFAR-100, and ImageNet32. We can observe that these open-set examples are difficult to disambiguate, i.e., their label confidence is not enough distinguishable. That is reasonable since they do not belong to these defined classes.  

\section{Conclusion}
In this paper, we pioneer a new PLL study that learns from examples whose true label is outside the candidate label set termed OOC (\underline{o}ut-\underline{o}f-\underline{c}andidate) examples. We consider two types of OOC examples in the real world including the closed-set (open-set) OOC examples whose true label is inside (outside) the known label space. To solve this new PLL problem, we propose a novel framework that first calculates the per-example wooden cross-entropy loss from candidate and non-candidate labels respectively, and dynamically differentiates the two types of OOC examples based on specially designed criteria. After that, we propose \emph{reversed label disambiguation} and \emph{random candidate generation} to further leverage selected closed-set and open-set OOC examples respectively for model training. Extensive experiments verify the superiority and effectiveness of the proposed method. In the future, it is interested to develop more effective ways for selecting OOC examples. Besides, since we have shown the positive role of open-set OOC examples, it is significant to better explore them for generalization. 

\section{Acknowledgement}
The authors wish to thank the anonymous reviewers for their helpful and valuable comments. Lei Feng is supported by the Joint NTU-WeBank Research Centre on Fintech (Award No: NWJ-2021-005), Nanyang Technological University, Singapore, the National Natural Science Foundation of China (Grant No. 62106028), Chongqing Overseas Chinese Entrepreneurship and Innovation Support Program, CAAI-Huawei MindSpore Open Fund, and Chongqing Artificial Intelligence Innovation Center. Guowu Yang is supported by the National Natural Science Foundation of China (Grant No. 62172075).

\bibliographystyle{ACM-Reference-Format}
\bibliography{reference}

\end{document}